\documentclass[conference]{IEEEtran}
\IEEEoverridecommandlockouts
\usepackage{cite}
\usepackage{amsmath,amssymb,amsfonts}
\usepackage{lipsum}
\usepackage{algorithmic}
\usepackage{graphicx}
\usepackage{textcomp}
\usepackage{xcolor}
\def\BibTeX{{\rm B\kern-.05em{\sc i\kern-.025em b}\kern-.08em
    T\kern-.1667em\lower.7ex\hbox{E}\kern-.125emX}}
\IEEEoverridecommandlockouts \IEEEpubid{\makebox[\columnwidth]{979-8-3503-8544-1/24/\$31.00~\copyright{}2024 IEEE \hfill} \hspace{\columnsep}\makebox[\columnwidth]{ }} 
\begin{document}

\title{CSI4Free: GAN-Augmented mmWave CSI for Improved Pose Classification\\
%{\footnotesize \textsuperscript{*}Note: Sub-titles are not captured in Xplore  and
%should not be used}

}

\author{\IEEEauthorblockN{Nabeel Nisar Bhat}
\IEEEauthorblockA{\textit{IDLab-Faculty of Science}\\
\textit{University of Antwerp-imec}\\
Antwerp, Belgium\\
nabeelnisar.bhat@uantwerpen.be
}
\and
\IEEEauthorblockN{Rafael Berkvens}
\IEEEauthorblockA{\textit{IDLab-Faculty of Applied Engineering}\\
\textit{University of Antwerp-imec}\\
Antwerp, Belgium\\
rafael.berkvens@uantwerpen.be
}
\and
\IEEEauthorblockN{Jeroen Famaey}
\IEEEauthorblockA{\textit{IDLab-Faculty of Science}\\
\textit{University of Antwerp-imec}\\
Antwerp, Belgium\\
jeroen.famaey@uantwerpen.be
}}

\maketitle 

\begin{abstract}
In recent years, Joint Communication and Sensing (JC\&S), has demonstrated significant success, particularly in utilizing sub-6 GHz frequencies with commercial-off-the-shelf (COTS) Wi-Fi devices for applications such as localization, gesture recognition, and pose classification. Deep learning and the existence of large public datasets has been pivotal in achieving such results. However, at mmWave frequencies (30-300 GHz), which has shown potential for more accurate sensing performance, there is a noticeable lack of research in the domain of COTS Wi-Fi sensing. Challenges such as limited research hardware, the absence of large datasets, limited functionality in COTS hardware, and the complexities of data collection present obstacles to a comprehensive exploration of this field. In this work, we aim to address these challenges by developing a method that can generate synthetic mmWave channel state information (CSI) samples. In particular, we use a generative adversarial network (GAN) on an existing dataset, to generate  30,000 additional CSI samples.  
The augmented samples exhibit a remarkable degree of consistency with the original data, as indicated by the notably high GAN-train and GAN-test scores. Furthermore, we integrate the augmented samples in training a pose classification model. We observe that the augmented samples complement the real data and improve the generalization of the classification model.

\end{abstract}


\section{Introduction}
In recent years, Wi-Fi signals have been widely utilized for sensing applications such as localization \cite{zhang20193d,wang2019joint}, gesture recognition\cite{wang2019joint,bhat2023gesture}, pose estimation \cite{bhat2023multi,jiang2020towards,wang2019can} and gait identification\cite{deng2022gaitfi}. In particular, channel state information (CSI) \cite{wang2019joint,bhat2023gesture,bhat2023multi,jiang2020towards,wang2019can} extracted from commercial-off-the-shelf (COTS) Wi-Fi access points (APs) has resulted in remarkable accuracy in these applications. The main advantage of using Wi-Fi signals for sensing is that most of the Wi-Fi infrastructure is already in place in homes, offices, and buildings. Therefore, the signals transmitted for communications can also be used for sensing, at a limited additional cost. This concept of using communication signals for sensing is known as joint communication and sensing (JC\&S) or integrated sensing and communication (ISAC) \cite{cui2021integrating}. Compared to camera-based sensing, Wi-Fi sensing offers improved privacy and does not require a well-lit environment \cite{zhang2019wigrus}. In Wi-Fi sensing, most of the focus has been on sub-6 GHz signals. However, researchers are now swiftly moving to mmWave (30-300 GHz) due to its large bandwidth and massive multiple input and multiple output (MIMO) capabilities. This has benefits not only limited to high data-rates but also more accurate sensing \cite{de2021convergent,struye2020towards}. Recently, Wi-Fi signals at mmWave have shown promising results in applications such as gesture recognition\cite{yu2020human,bhat2023gesture}, pose estimation\cite{bhat2023multi}, and localization\cite{koike2020fingerprinting}. In this work, we focus on mmWave and pose classification. \\The major force behind the exceptional success of Wi-Fi sensing is deep learning\cite{wang2019joint,bhat2023gesture,bhat2023multi,jiang2020towards}. Deep learning has achieved state-of-the-art performance on many Wi-Fi sensing tasks. One of the key factors contributing to the effectiveness of deep learning methods, in general, is the availability of extensive and varied datasets. Unfortunately, Wi-Fi sensing at mmWave frequencies has suffered due to a lack of research hardware and consequently, the non-existence of large datasets. Moreover, data collection and annotation with Wi-Fi is difficult and time-consuming. Also, the lack of visual cues or ground truth in the Wi-Fi signal does not help either. To avoid overfitting problems, deep learning requires a large amount of labeled data\cite{antoniou2018augmenting}.
\\ In this work, we aim to reduce the effort required in the data collection by employing data augmentation techniques. Generative adversarial networks (GANs) \cite{goodfellow2014generative} have proven to generate realistic and high-quality synthetic or fake images. GANs have been used for image augmentation \cite{frid2018gan}, anomaly detection\cite{li2019mad}, super-resolution\cite{you2019ct}, 3D object generation \cite{smith2017improved} and domain adaptation\cite{tzeng2017adversarial}. Our focus is on augmentation. Augmenting CSI is difficult and somewhat limited in literature as standard augmentation methods used for images such as random crop, horizontal flip, rotation, and brightness can not be directly applied to CSI data. Moreover, these methods produce a limited set of augmentations \cite{antoniou2018augmenting}. Differently, GAN-based augmentation methods have huge potential \cite{bhattacharya2020gan,antoniou2018augmenting} and can be used to generate more natural and a broad set of augmentations. GANs mimic the original distribution of the dataset to generate more realistic samples than the standard augmentation methods and also improve the generalization of a downstream model. However, GANs are difficult to train and often suffer from what is known as mode collapse\cite{thanh2020catastrophic}. Moreover, GAN training is domain specific. Therefore, the utilization of GANs in augmenting CSI data is limited \cite{han2020deep}, especially in the context of COTS mmWave CSI data where their potential has not yet been explored. \\ In this work, we propose a GAN-based method for augmenting mmWave CSI data. Specifically, we use a conditional GAN (cGAN) \cite{mirza2014conditional} to generate new synthetic samples. We train our method on an existing mmWave COTS dataset \cite{bhat2023multi}, consisting of 3 users performing a set of 8 distinct poses. We carefully train the generator and discriminator of cGAN and generate an additional 30,000 synthetic CSI samples, approximately 3800 samples for each pose. In this way, we increase the sample size of the dataset from 1084 to 31184 samples. We then validate the consistency of GAN-generated samples using GAN-train and GAN-test scores\cite{shmelkov2018good}. Our results reveal high GAN-train and GAN-test scores indicating a high quality of synthetic samples. Finally, we show that the synthetic samples also improve the performance of the downstream model for pose classification. Through this method, we create a fairly large mmWave CSI dataset which can be used by researchers to test and validate complex signal processing and deep learning models for pose classification. 

\section{Related Work} \label{related work}
\subsection{COTS mmWave Sensing}
Wi-Fi signals at sub-6 GHz have shown great potential in sensing applications. However, range resolution and spatial resolution at these frequencies are limited due to limited available bandwidth. Instead at mmWave and THz frequencies, larger bandwidths are available, potentially leading to sub-cm-level localization and high-definition imaging \cite{de2021convergent}. In this section, we review the recent developments in mmWave sensing, mainly focusing on COTS Wi-Fi.
\\ Yu et al. \cite{yu2020human} made pioneering contributions in the field of mmWave Wi-Fi sensing. The authors used mid-grained spatial beam signal-to-noise ratios (SNRs) for human pose and seat occupancy detection. The testbed consisted of a pair of Talon AD 7200 COTS routers. An open-source tool \cite{steinmetzer2017talon} was used to gain access to beam SNR channel measurements from the routers. 8 distinct poses were performed. The authors used deep learning-based methodology and achieved an overall accuracy of 90\% for pose classification. However, the dataset involves a single person and is not publicly available. In our previous work \cite{bhat2023gesture}, we validated the potential of mmWave Wi-Fi sensing for gesture recognition. We collected beam SNRs and CSI from mmWave and sub-6 GHz Wi-Fi APs respectively and compared the performance of the two approaches for gesture recognition. The dataset consisted of 10 distinct but closely related gestures/poses across 3 people and 2 environments. We achieved 96.7\% accuracy with mmWave Wi-Fi for gesture recognition in a single environment. However, incorporating more users and environments leads to a reduction of accuracy by 6\%. This is because of the limited dataset owing to the fact that data collection with mmWave COTS Wi-Fi is labor-intensive. Moreover, the low sample rate of beam SNRs (10 samples per second) adds to the complexity of data collection. Pegoraro et al \cite{pegoraro2023disc} recently published a dataset for integrated sensing and communication based on a software-defined ratio. The authors highlight the lack of research hardware at mmWave and consequently lack of datasets. The testbed is capable of transmitting 60 GHz IEEE 802-11-ay packets. The dataset consists of 7 subjects performing 4 activities walking, running, standing up-sitting down, and waving hands. However, the testbed is difficult to replicate due to cost and complexity. Moreover, limited activities are performed.
More recently, we in our work \cite{bhat2023multi}, for the first time used CSI from COTS mmWave Wi-Fi AP for pose estimation (regression) and pose classification. The testbed consisted of MikroTik wAP 60Gx3 COTS routers. We followed the work \cite{blanco2022augmenting} and installed OpenWrt to get access to CSI measurements from the routers. We employed a deep neural network-based methodology to derive a full-body pose from mmWave Wi-Fi. Microsoft Kinect was used to record the ground truth. The validation was performed across 3 individuals with 8 distinct poses. We achieved a high pose classification accuracy ($>$90\%) for the classification task and a low mean square error (MSE) of 0.0048 for the regression task (full body pose). 
Also, in this case, the sampling rate of CSI was limited to around 22 samples per second, resulting in a limited dataset size. Note that it is not possible to increase sampling frequency in these devices as the firmware is not efficient, leading to stability issues. Moreover, increasing the sampling frequency of CSI arbitrarily, implies misemploying the concept of JC\&S.  
\\The above discussion validates the use of mmWave Wi-Fi for sensing. Nevertheless, the progress in COTS mmWave research has been hindered by factors such as the absence of research-grade hardware, hardware limitations, labour-intensive data collection, and the scarcity of publicly available datasets. In this work, our goal is to minimize the challenges associated with data collection and to create an extensive dataset for pose classification, building upon our earlier research \cite{bhat2023multi}. To achieve this goal, we leverage generative adversarial networks (GANs) for data augmentation. 
\subsection{Data Augmentation}
\begin{figure*} [!t]
    \centering     \includegraphics[width=13.5 cm,]{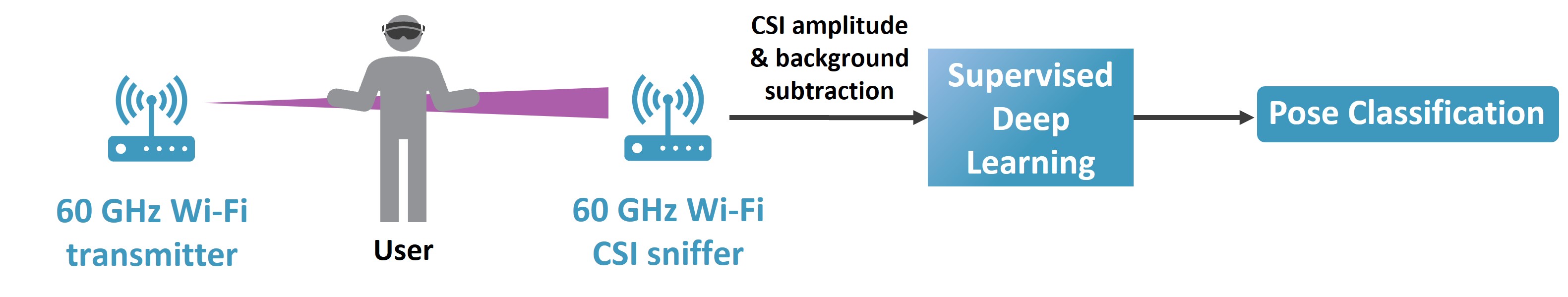}
    \caption{\textbf{JC\&S pose estimation framework}. A user performs a set of poses in between 60 GHz Wi-Fi devices, creating unique patterns in channel state information (CSI). A CSI sniffer captures the CSI. The amplitude of the CSI is fed to a deep learning-based model to map changes in CSI to different poses.}
    \label{fig:losses}
\end{figure*} 
Efficient training of neural networks requires a huge amount of data \cite{bhattacharya2020gan}. When the dataset is limited, network parameters are often undetermined and generalize poorly. To combat this, data augmentation can be considered. However, standard data augmentation methods produce limited plausible data \cite{antoniou2018augmenting}. Waheed et al. \cite{waheed2020covidgan} proposed CovidGAN in their work, aiming to generate synthetic chest X-ray images for enhancing COVID-19 detection. Their findings indicate a notable 10\% improvement in classification accuracy through the integration of synthetic images in the downstream task. Bhattacharya et al. \cite{bhattacharya2020gan} used Deep Convolutional Generative Adversarial Network (DCGAN) for data augmentation, to combat class imbalance in medical datasets. With DCGAN-augmented data, the classification accuracy rose from 60\% to 65\%. Han et al. \cite{han2020deep} used GAN for CSI data augmentation to reduce the effort of data collection and prevent overfitting risks caused by incomplete datasets. The authors also evaluated the quality of GAN-generated samples. The experiments were conducted at sub-6 GHz frequencies with Wi-Fi signals. The authors went a step further towards domain adaptation and compared performances with and without data augmentation as the first step. In the former case, the authors observed around 7\% accuracy in the target (unseen environment) domain. However, only four gestures were considered and all were operated by the right hand.
\\ Although GANs have been effectively employed for tasks involving CSI data at sub-6 GHz frequencies \cite{zou2018robust}, their application to COTS mmWave data remains undocumented at present.
\section{Methodology}
\subsection{JC\&S Pose Classification pipeline} \label{subsection:JCS}
Our pose classification pipeline consists of 60 GHz COTS Wi-Fi devices. In particular, we use MikroTik wAP 60x3 Wi-Fi devices, one acting as a transmitter and the other as a CSI sniffer. We follow the work of Blanco
et al. \cite{blanco2022augmenting} to gain access to CSI measurements. In our setup, a user performs a set of poses between the two devices. This creates unique patterns in CSI captured by the sniffer. The amplitude details of the CSI are subsequently input into a convolutional neural network (CNN), to extract distinctive features and effectively map changes in CSI to distinct poses \cite{bhat2023multi}. However, in our approach, we opt to synthetically generate CSI to improve the generalization of the deep learning model for pose classification.
\subsection{Generative Adversarial Networks (GANs)}
Generative adversarial networks (GANs) \cite{goodfellow2014generative} aim to generate new data that mirrors the statistical characteristics of the training data. GANs are very good at modeling high-dimensional distributions of the data. From the learned distribution, new data can be generated visually indistinguishable from the real data. GANs belong to a class of generative models that try to learn realistic representations of a class. These models take random noise ($z$) as an input and sometimes, a class label (Y). From the input, generative models generate a set of features (X) that resemble a particular class. $z$ ensures there is diversity in X generated by the model. Therefore, generative models try to capture conditional probability  P(X$\mid$Y). On the other hand, discriminative models often called classifiers, try to distinguish between the classes. These models take a set of X as input and determine the corresponding Y. Discriminative models aim to learn P(Y$\mid$X). 
\\ GANs consist of two neural networks, Generator ($\mathcal{G}$) and Discriminator ($\mathcal{D}$), trained alternatively and competing with each other. $\mathcal{G}$ takes $z$ as an input usually sampled from a normal distribution, $z \sim \mathcal{N}(\mu, \sigma$),
 and tries to generate fake or synthetic data ($\mathcal{G}(z)$) to fool $\mathcal{D}$. Note that the terms fake, generated, or synthetic are used interchangeably to describe the data produced by $\mathcal{G}$.
 $z$ is often a low-dimensional vector and the corresponding sample space is called latent space. $\mathcal{G}(z)$ is a high-dimensional fake output e.g., an RGB image or audio or CSI, in our case. The goal of $\mathcal{D}$ is to correctly classify real and fake data. So, $\mathcal{D}$ outputs a single probability of an input being real or fake. These probabilities are fed back to the $\mathcal{G}$ to improve its output. GANs are trained in an unsupervised manner or indirect training, in the sense that $\mathcal{G}$ does not get to see real samples or images, but is trained to fool $\mathcal{D}$. The optimization of GANs is a two-player min-max optimization problem that terminates at a saddle point forming a minimum with respect to $\mathcal{G}$ and maximum with respect to $\mathcal{D}$ \cite{gui2021review}. The goal of the optimization is to reach Nash equilibrium \cite{ratliff2013characterization}, a point where no player can improve by changing its weights. At this point, $\mathcal{G}$ can be considered to have captured the distribution of real samples. The optimization can be mathematically formulated as follows:
\begin{equation}  \label{eq:BCE}
\begin{aligned}
&\min_\mathcal{G} \max_\mathcal{D} V(\mathcal{D}, \mathcal{G}) = \mathbb{E}_{x \sim p_{\text{data}}(x)}\left[\log \mathcal{D}(x)\right] \\
&+ \mathbb{E}_{z \sim p_z(z)}\left[\log(1 - \mathcal{D}(\mathcal{G}(z)))\right]
\end{aligned}
\end{equation}
where $V(\mathcal{D},\mathcal{G})$ is the reward, $x$ is the real data (real CSI), $\mathcal{D}(x)$ and $\mathcal{D}(\mathcal{G}(z))$ represent output of $\mathcal{D}$ on real CSI and GAN-generated synthetic data (synthetic CSI), respectively. $\mathbb{E}_x$ is the expected value over all real CSI instances. $\mathbb{E}_z$ is the expected value over all random inputs to the generator. The formula derives from the cross-entropy between the real and generated distributions. $\mathcal{D}$ has access to both x and $\mathcal{G}(z)$. An ideal $\mathcal{D}$ would output 1 for x and 0 for $\mathcal{G}(z)$ i.e., classifying real as real and synthetic as synthetic. On the other hand, $\mathcal{G}$ only sees synthetic CSI, and aims to push $\mathcal{D}(\mathcal{G}(z))$ close to 1, to minimize the overall loss function. This optimization problem is implemented as Binary Cross Entropy (BCE) loss.
\subsection{Method: Conditional Wasserstein GAN (cWGAN)} \label{subsection:cwgan}
\begin{figure*} [!t]
    \centering     \includegraphics[width=13.5 cm,scale=1]{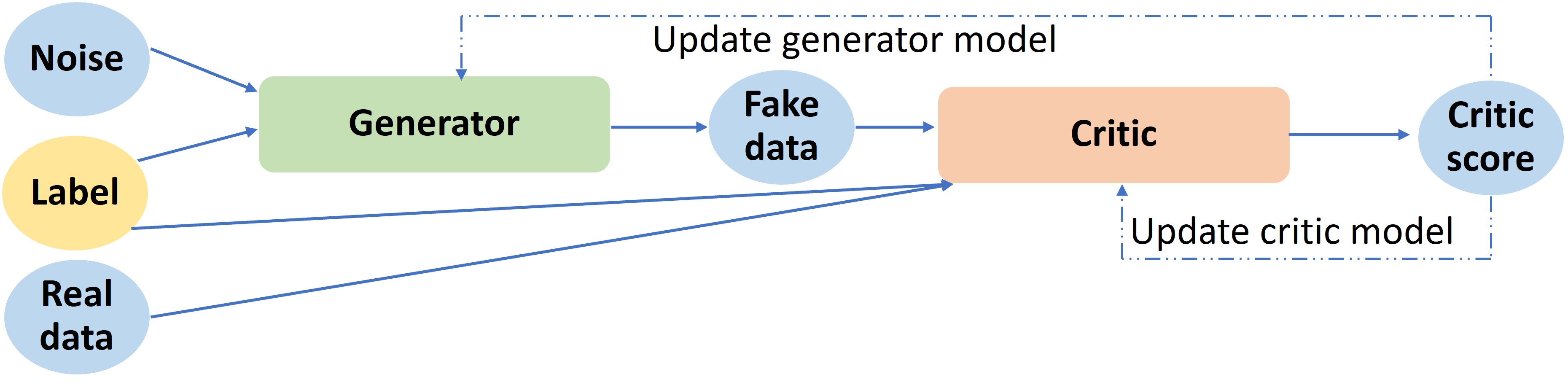}
    \caption{\textbf{Our GAN-based augmentation framework}. We propose an augmentation method for mmWave CSI based on cWGAN. The generator generates high-dimensional synthetic CSI by sampling low-dimensional noise, with the goal of minimizing the distance between the synthetic CSI and the real CSI. While the critic tries to separate the two distributions apart.  Based on the critic's feedback, the generator is encouraged to produce more realistic CSI. }
    \label{fig:cGAN}
\end{figure*}
We use our previously collected dataset \cite{bhat2023gesture}  consisting of CSI samples corresponding to 3 users, performing a set of 8 distinct poses (8 classes). We use only amplitude information of the CSI, as phase information is noisy and calibration is needed. For each user, x $\in \mathbb{R}^{m \times n \times p }$. x represents real CSI, m represents the number of samples/examples, and n and p represent the antenna and time index, respectively. We propose a method based on conditional GAN (cGAN) for data augmentation of the CSI samples. Since cGANs can be conditioned on labels, we can control the generation process, unlike standard GANs. cGANs therefore learn mapping from input to output (classes) and achieve faster convergence. cGANs have an additional input layer of one-hot-encoded labels (label embedding layer). This additional layer guides the generator in terms of which image or sample to produce. To train our cGAN, we first use a BCE loss according to Equation \ref{eq:BCE}. However, we observe mode collapse and unstable training with BCE loss. To counter this,
we adopt an enhanced loss function known as the Wasserstein Loss (W-loss) \cite{gulrajani2017improved}. This loss function offers improved stability in the training of GANs. W-loss is implemented in the following way:
\begin{equation} \label{eq:Wloss}
\begin{aligned}
\min_\mathcal{G} \max_\mathcal{C} \left(\mathbb{E}_{x \sim p_{\text{data}}}\left[\mathcal{C}(x)\right] - \mathbb{E}_{z \sim p_z}\left[\mathcal{C}(\mathcal{G}(z))\right]\right)
\end{aligned}
\end{equation}

Figure \ref{fig:cGAN} shows the architecture of cGAN with W-loss (cWGAN). Here, the discriminator is called critic ($\mathcal{C}$), since its output can be any real number, not bounded between 0 and 1. $G$ tries to minimize the loss by maximizing $\mathcal{C}(G(z))$, bringing the real distribution closer to fake one. Instead, the aim of $\mathcal{C}$ is to maximize the distance and separate the two distributions apart. For W-loss to be valid and approximate Earth Mover's Distance (EMD), the critic's neural network needs to be 1-Lipschitz  continuous, which means the norm of its gradients should be at most 1. This ensures that W-loss is valid and does not grow much. We encourage this, by adding a gradient penalty \cite{gulrajani2017improved} as follows:
\begin{equation} \label{eq:wlossgp}
\begin{aligned}
&\min_\mathcal{G} \max_\mathcal{C} (\mathbb{E}_{x \sim p_{\text{data}}}\left[\mathcal{C}(x)\right] - \mathbb{E}_{z \sim p_z}\left[\mathcal{C}(\mathcal{G}(z))\right] 
\\&+ \lambda \mathbb{E}_{\hat{x} \sim p_{\hat{x}}}\left[(\|\nabla_{\hat{x}} \mathcal{C}(\hat{x})\|_2 - 1)^2\right])
\end{aligned}
\end{equation}
where \(\hat{x} = \epsilon x + (1 - \epsilon) \mathcal{G}(z)\).\\
$\hat{x}$ is interpolation between real and synthetic CSI, weighted by $\epsilon$, \(\nabla\) represents the gradient operator, \(\| \cdot \|_2\) represents the squared Euclidean norm and \(\lambda\) controls magnitude of gradient penalty. We use a linear $\mathcal{G}$ and a linear $\mathcal{C}$. $\mathcal{G}$ consists of a label embedding layer and five linear layers. The linear layers except the last one, are followed by LeakyReLU activations. Tanh is used after the last linear layer to scale inputs between -1 and 1. $\mathcal{C}$ consists of a label embedding layer and 3 linear layers. LeakReLU activation is employed after the first two linear layers. Additionally, Dropout is used after the second linear layer. No activation is used after the last linear layer for W-loss to work. The $\mathcal{C}$ outputs a single real value, its score on real and fake samples. We use cWGAN to generate 30,000 samples of CSI for each user, without any additional data collection\footnote{Dataset is available at:https://zenodo.org/records/10702215}. However, the challenge is to evaluate the quality of generated samples as the CSI samples lack visual information unlike images, where one can visually inspect the quality of samples. To combat this, we adopt GAN-train and GAN-test metrics presented in the work \cite{shmelkov2018good}. GAN-train involves training a classification model on data generated by a GAN and evaluating it on real data. If the model, which exclusively encounters GAN-generated data during training, achieves high accuracy on real data, it suggests that the generated samples closely resemble real ones. A high GAN-train score indicates diversity in the generated samples. On the other hand, the GAN-test assesses the accuracy of a model trained on real data and evaluated on GAN-generated data. A high score implies that GAN-generated samples realistically approximate the (unknown) distribution of real samples. To achieve this, we use a supervised deep learning model based on CNN (cf., Figure \ref{fig:losses}), which serves as a downstream model for pose classification. Our CNN consists of 3 convolution layers, each followed by batch norm, ReLU, and pooling layers. These layers extract features from the input data and encode the high-dimensional information into a low-dimensional space. Finally, a linear layer is employed to output a score for 8 classes. As an additional validation step, we also evaluate if the GAN-generated samples improve the classification accuracy of the original task (pose classification).
\section{Experiments}
\subsection{GAN Training}
First, we use BCE loss with cGAN. We use a linear $\mathcal{G}$ and a linear $\mathcal{D}$. Specifically, we use 5 linear layers for $\mathcal{G}$ and $\mathcal{D}$. Normalization layers are used for both $\mathcal{G}$ and $\mathcal{D}$. Besides, ReLU activation is used for $\mathcal{G}$ and LeakyReLU for $\mathcal{D}$. This architectural setup draws inspiration from the design commonly used in computer vision tasks and best practices for GANs. The sigmoid layer is used as the final layer of $\mathcal{D}$ and tanh for the $\mathcal{G}$. We sample the input noise from a standard normal distribution and set the dimensionality (latent space) of the noise vector to 100. We use Adam as an optimizer with a learning rate of 3e-4. Batch size is set to 32 and weights of $\mathcal{G}$ and $\mathcal{D}$ are initialized to normal distribution. $\mathcal{G}$ takes noise vector and labels as an input and outputs $m\times1500$ where $m$ is the number of samples of synthetic CSI and 1500 ($30\times50$, representing antenna index and time index respectively). On the other hand, $\mathcal{D}$ takes CSI (synthetic or real) and corresponding labels as input and predicts the probability of the CSI sample being real or synthetic. 
We train cGAN for 30,000 epochs and after every 500 epochs, we extract and save the synthetic CSI and corresponding labels. Besides, we monitor the $\mathcal{G}$ loss, $\mathcal{D}$ loss, and $\mathcal{D}$ accuracy to see if the training is stable. Additionally, we use our evaluation metric (GAN-train and GAN-test), to inspect the quality of generated samples. We see that with BCE loss, the training process does not converge and $\mathcal{G}$ loss increases steadily. Moreover, $\mathcal{D}$ accuracy quickly approaches 100\% suggesting that $\mathcal{G}$ is not able to fool $\mathcal{D}$. Further, the GAN-train and GAN-test scores are significantly low for all 3 users, well below 30\%.
\begin{figure} [!t]
    \centering     \includegraphics[width=7 cm,scale=1]{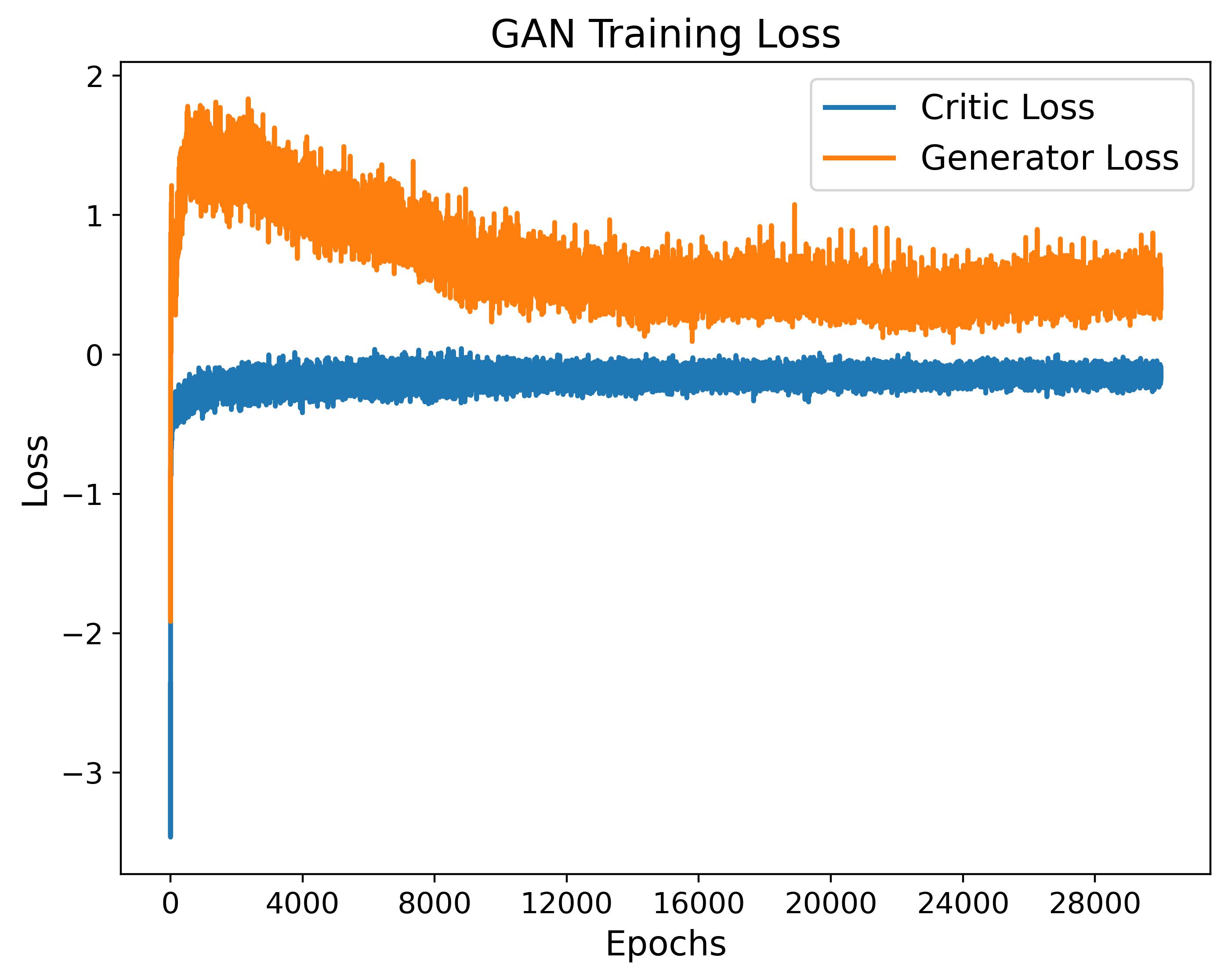}
    \caption{Training cGAN using W-loss with linear $\mathcal{G}$ and linear $\mathcal{C}$.}
    \label{fig:cwgan}
\end{figure} 

\begin{table}[!t]
\caption{Validation of cGAN with W-loss.}
\begin{center}
\begin{tabular}{ c c c c c} 
\hline
User$\#$ & GAN-train    & GAN-test  & Baseline Acc.& cWGAN Acc.
\\ \hline
1    & 98.8\%   & 95.5\%  & 96.4\% & 99.7\% \\
2     & 99.6\%   & 96\%  & 96.18\% & 99.3\% \\
3    & 88\%   & 91\%  & 90.05\% & 94.1\% \\
\hline
\end{tabular}
\label{table:table2}
\end{center}
\end{table}
Next, we try to make $\mathcal{G}$ more powerful by using convolution layers in the architecture. However, also in this case, we observe low GAN-train and GAN-test scores. We link this failure to the mode collapse of GANs due to BCE loss. Therefore, this approach with BCE loss does not work.
\subsection{cWGAN}
Due to the above problems, we adopt W-loss instead of BCE loss and further tune the GAN training process. We use the $\mathcal{G}$ and $\mathcal{C}$ described in Section \ref{subsection:cwgan}. In this case, we do not use any normalization layers. Using normalization layers leads to unstable training. We use the same set of hyper-parameters as mentioned previously. In addition, we stick to the default value of 10 for $\lambda$, which controls the magnitude of the gradient penalty in Equation \ref{eq:wlossgp}. Moreover, we train $\mathcal{G}$ once for every 5 iterations of $\mathcal{C}$. In other words, we allow $\mathcal{C}$ to be strong. This encourages $\mathcal{G}$ to be updated with better gradients and adds stability to the overall process.
Figure \ref{fig:cwgan} shows the training process of cWGAN with linear $\mathcal{G}$ and linear $\mathcal{C}$ for user 1. From Figure \ref{fig:cwgan}, it is quite evident that the losses are bounded and GAN training converges. The same holds for other users. We use the $\mathcal{G}$ of trained cWGAN to generate 30,000 synthetic samples of CSI for each user and compute GAN-train and GAN-test scores for these samples, as mentioned in Section \ref{subsection:cwgan}. Moreover, we also evaluate if the synthetic samples lead to an increase in the task of pose classification. Table \ref{table:table2} shows the performance of our method on 3 different users (persons). GAN-train and GAN-test scores have been introduced in Section \ref{subsection:cwgan}, Baseline Acc. refers to the original pose classification accuracy with the real CSI. This is obtained by splitting real CSI into train and test splits. We use a standard 75:25 split for train and test sets, respectively. Then a classifier, CNN, described in Section \ref{subsection:cwgan} is trained on train split and evaluated on test split. Instead, cWGAN Acc. refers to the pose classification accuracy when the same classifier is trained on train splits of real CSI and GAN-generated CSI and evaluated on the test split of the real CSI. In other words, the latter measures the impact of augmentation on the original task. 
One can clearly see that cWGAN-generated samples get a very high GAN-train and GAN-test score, similar to typical validation accuracy (Baseline Acc.). Thus, the synthetic samples are highly consistent with the actual data. In this way, we not only increase the size of the dataset but also maintain the consistency of the samples. Further, we can see that in all three cases, we get an improved performance in pose classification accuracy, when generated data is combined with the actual data. We notice 3.3\%, 3.1\%, and 4\% improvement in accuracy for users 1, 2, and 3 respectively, compared to Baseline Acc. This high pose classification accuracy is crucial when deploying mmWave ISAC-based solutions for real-world applications. 
\section{Conclusion and Future Work}
In this study, we successfully conducted stable training of GANs for mmWave CSI. Notably, we adopted a cWGAN-based approach to achieve robust data augmentation. However, training the cGAN with BCE loss proved ineffective as the synthetic samples exhibited low scores in both GAN-train and GAN-test evaluations. Consequently, we adopted cWGAN, cGAN with W-loss. This modification resulted in high-quality synthetic CSI with high validation scores as far as GAN-train and GAN-test metrics are concerned. The result is a large dataset of COTS mmWave Wi-Fi samples that can be used by researchers for validating their signal processing and deep learning methodology for pose classification. Further, we also showed that cWGAN-generated data complements the real data in the original task of pose classification, resulting in improved generalization. Our method represents an initial stride towards domain adaptation. In the future, our goal will be to achieve broader generalization across different people and environments with mmWave Wi-Fi leveraging GANs. Additionally, we will also explore the transferability of our method to other related tasks beyond pose classification.
\section*{Acknowledgment}
This research is funded by the FWO project (Grant number: 1SH5X24N) and  FWO WaveVR (Grant number: G034322N).
\bibliographystyle{IEEEbib}
\bibliography{references}

\begin{thebibliography}{10}

\bibitem{zhang20193d}
Lingyan Zhang and Hongyu Wang,
\newblock ``{3D-Wi-Fi}: 3{D} localization with commodity {Wi-Fi},''
\newblock {\em IEEE Sensors Journal}, vol. 19, no. 13, pp. 5141--5152, 2019.

\bibitem{wang2019joint}
Fei Wang, Jianwei Feng, Yinliang Zhao, Xiaobin Zhang, Shiyuan Zhang, and Jinsong Han,
\newblock ``Joint activity recognition and indoor localization with {Wi-Fi} fingerprints,''
\newblock {\em IEEE Access}, vol. 7, pp. 80058--80068, 2019.

\bibitem{bhat2023gesture}
Nabeel~Nisar Bhat, Rafael Berkvens, and Jeroen Famaey,
\newblock ``Gesture recognition with {mmWave} {Wi-Fi} access points: Lessons learned,''
\newblock in {\em 2023 IEEE 24th International Symposium on a World of Wireless, Mobile and Multimedia Networks (WoWMoM)}. IEEE, 2023, pp. 127--136.

\bibitem{bhat2023multi}
Nabeel~Nisar Bhat, Javad Sameri, Jakob Struye, Maria~Torres Vega, Rafael Berkvens, and Jeroen Famaey,
\newblock ``Multi-modal pose estimation in {XR} applications leveraging integrated sensing and communication,''
\newblock in {\em Proceedings of the 1st ACM Workshop on Mobile Immersive Computing, Networking, and Systems}, 2023, pp. 261--267.

\bibitem{jiang2020towards}
Wenjun Jiang, Hongfei Xue, Chenglin Miao, Shiyang Wang, Sen Lin, Chong Tian, Srinivasan Murali, Haochen Hu, Zhi Sun, and Lu~Su,
\newblock ``Towards 3{D} human pose construction using {Wi-Fi},''
\newblock in {\em Proceedings of the 26th Annual International Conference on Mobile Computing and Networking}, 2020, pp. 1--14.

\bibitem{wang2019can}
Fei Wang, Stanislav Panev, Ziyi Dai, Jinsong Han, and Dong Huang,
\newblock ``Can {Wi-Fi} estimate person pose?,''
\newblock {\em arXiv preprint arXiv:1904.00277}, 2019.

\bibitem{deng2022gaitfi}
Lang Deng, Jianfei Yang, Shenghai Yuan, Han Zou, Chris~Xiaoxuan Lu, and Lihua Xie,
\newblock ``Gaitfi: Robust device-free human identification via {Wi-Fi} and vision multimodal learning,''
\newblock {\em IEEE Internet of Things Journal}, vol. 10, no. 1, pp. 625--636, 2022.

\bibitem{cui2021integrating}
Yuanhao Cui, Fan Liu, Xiaojun Jing, and Junsheng Mu,
\newblock ``Integrating sensing and communications for ubiquitous {IoT}: Applications, trends, and challenges,''
\newblock {\em IEEE Network}, vol. 35, no. 5, pp. 158--167, 2021.

\bibitem{zhang2019wigrus}
Tao Zhang, Tingyu Song, Daolin Chen, Tian Zhang, and Jie Zhuang,
\newblock ``Wigrus: A {Wi-Fi}-based gesture recognition system using software-defined radio,''
\newblock {\em IEEE Access}, vol. 7, pp. 131102--131113, 2019.

\bibitem{de2021convergent}
Carlos De~Lima, Didier Belot, Rafael Berkvens, Andre Bourdoux, Davide Dardari, Maxime Guillaud, Minna Isomursu, Elena-Simona Lohan, Yang Miao, Andre~Noll Barreto, et~al.,
\newblock ``Convergent communication, sensing and localization in {6G} systems: An overview of technologies, opportunities and challenges,''
\newblock {\em IEEE Access}, vol. 9, pp. 26902--26925, 2021.

\bibitem{struye2020towards}
Jakob Struye, Filip Lemic, and Jeroen Famaey,
\newblock ``Towards ultra-low-latency {mmWave} {Wi-Fi} for multi-user interactive virtual reality,''
\newblock in {\em GLOBECOM 2020-2020 IEEE Global Communications Conference}. IEEE, 2020, pp. 1--6.

\bibitem{yu2020human}
Jianyuan Yu, Pu~Wang, Toshiaki Koike-Akino, Ye~Wang, Philip~V Orlik, and Haijian Sun,
\newblock ``Human pose and seat occupancy classification with commercial {mmWave} {Wi-Fi},''
\newblock in {\em 2020 IEEE Globecom Workshops (GC Wkshps}. IEEE, 2020, pp. 1--6.

\bibitem{koike2020fingerprinting}
Toshiaki Koike-Akino, Pu~Wang, Milutin Pajovic, Haijian Sun, and Philip~V Orlik,
\newblock ``Fingerprinting-based indoor localization with commercial {mmWave} {Wi-Fi}: A deep learning approach,''
\newblock {\em IEEE Access}, vol. 8, pp. 84879--84892, 2020.

\bibitem{antoniou2018augmenting}
Antreas Antoniou, Amos Storkey, and Harrison Edwards,
\newblock ``Augmenting image classifiers using data augmentation generative adversarial networks,''
\newblock in {\em Artificial Neural Networks and Machine Learning--ICANN 2018: 27th International Conference on Artificial Neural Networks, Rhodes, Greece, October 4-7, 2018, Proceedings, Part III 27}. Springer, 2018, pp. 594--603.

\bibitem{goodfellow2014generative}
Ian Goodfellow, Jean Pouget-Abadie, Mehdi Mirza, Bing Xu, David Warde-Farley, Sherjil Ozair, Aaron Courville, and Yoshua Bengio,
\newblock ``Generative adversarial nets,''
\newblock {\em Advances in neural information processing systems}, vol. 27, 2014.

\bibitem{frid2018gan}
Maayan Frid-Adar, Idit Diamant, Eyal Klang, Michal Amitai, Jacob Goldberger, and Hayit Greenspan,
\newblock ``{GAN}-based synthetic medical image augmentation for increased cnn performance in liver lesion classification,''
\newblock {\em Neurocomputing}, vol. 321, pp. 321--331, 2018.

\bibitem{li2019mad}
Dan Li, Dacheng Chen, Baihong Jin, Lei Shi, Jonathan Goh, and See-Kiong Ng,
\newblock ``{MAD-GAN}: Multivariate anomaly detection for time series data with generative adversarial networks,''
\newblock in {\em International conference on artificial neural networks}. Springer, 2019, pp. 703--716.

\bibitem{you2019ct}
Chenyu You, Guang Li, Yi~Zhang, Xiaoliu Zhang, Hongming Shan, Mengzhou Li, Shenghong Ju, Zhen Zhao, Zhuiyang Zhang, Wenxiang Cong, et~al.,
\newblock ``{CT} super-resolution {GAN} constrained by the identical, residual, and cycle learning ensemble (gan-circle),''
\newblock {\em IEEE transactions on medical imaging}, vol. 39, no. 1, pp. 188--203, 2019.

\bibitem{smith2017improved}
Edward~J Smith and David Meger,
\newblock ``Improved adversarial systems for 3{D} object generation and reconstruction,''
\newblock in {\em Conference on Robot Learning}. PMLR, 2017, pp. 87--96.

\bibitem{tzeng2017adversarial}
Eric Tzeng, Judy Hoffman, Kate Saenko, and Trevor Darrell,
\newblock ``Adversarial discriminative domain adaptation,''
\newblock in {\em Proceedings of the IEEE conference on computer vision and pattern recognition}, 2017, pp. 7167--7176.

\bibitem{bhattacharya2020gan}
Debangshu Bhattacharya, Subhashis Banerjee, Shubham Bhattacharya, B~Uma~Shankar, and Sushmita Mitra,
\newblock ``{GAN-based} novel approach for data augmentation with improved disease classification,''
\newblock {\em Advancement of Machine Intelligence in Interactive Medical Image Analysis}, pp. 229--239, 2020.

\bibitem{thanh2020catastrophic}
Hoang Thanh-Tung and Truyen Tran,
\newblock ``Catastrophic forgetting and mode collapse in {GANs},''
\newblock in {\em 2020 international joint conference on neural networks (ijcnn)}. IEEE, 2020, pp. 1--10.

\bibitem{han2020deep}
Zijun Han, Lingchao Guo, Zhaoming Lu, Xiangming Wen, and Wei Zheng,
\newblock ``Deep adaptation networks based gesture recognition using commodity {Wi-Fi},''
\newblock in {\em 2020 IEEE Wireless Communications and Networking Conference (WCNC)}. IEEE, 2020, pp. 1--7.

\bibitem{mirza2014conditional}
Mehdi Mirza and Simon Osindero,
\newblock ``Conditional generative adversarial nets,''
\newblock {\em arXiv preprint arXiv:1411.1784}, 2014.

\bibitem{shmelkov2018good}
Konstantin Shmelkov, Cordelia Schmid, and Karteek Alahari,
\newblock ``How good is my {GAN}?,''
\newblock in {\em Proceedings of the European conference on computer vision (ECCV)}, 2018, pp. 213--229.

\bibitem{steinmetzer2017talon}
Daniel Steinmetzer, Daniel Wegemer, and Matthias Hollick,
\newblock ``Talon tools: The framework for practical ieee 802.11 ad research.(2017),'' 2017.

\bibitem{pegoraro2023disc}
Jacopo Pegoraro, Jesus~Omar Lacruz, Michele Rossi, and Joerg Widmer,
\newblock ``Disc: a dataset for integrated sensing and communication in {mmWave} systems,''
\newblock {\em arXiv preprint arXiv:2306.09469}, 2023.

\bibitem{blanco2022augmenting}
Alejandro Blanco, Pablo~Jim{\'e}nez Mateo, Francesco Gringoli, and Joerg Widmer,
\newblock ``Augmenting {mmWave} localization accuracy through sub-6 ghz on off-the-shelf devices,''
\newblock in {\em Proceedings of the 20th Annual International Conference on Mobile Systems, Applications and Services}, 2022, pp. 477--490.

\bibitem{waheed2020covidgan}
Abdul Waheed, Muskan Goyal, Deepak Gupta, Ashish Khanna, Fadi Al-Turjman, and Pl{\'a}cido~Rogerio Pinheiro,
\newblock ``Covid{GAN}: data augmentation using auxiliary classifier {GAN} for improved covid-19 detection,''
\newblock {\em Ieee Access}, vol. 8, pp. 91916--91923, 2020.

\bibitem{zou2018robust}
Han Zou, Jianfei Yang, Yuxun Zhou, Lihua Xie, and Costas~J Spanos,
\newblock ``Robust {Wi-Fi}-enabled device-free gesture recognition via unsupervised adversarial domain adaptation,''
\newblock in {\em 2018 27th International Conference on Computer Communication and Networks (ICCCN)}. IEEE, 2018, pp. 1--8.

\bibitem{gui2021review}
Jie Gui, Zhenan Sun, Yonggang Wen, Dacheng Tao, and Jieping Ye,
\newblock ``A review on generative adversarial networks: Algorithms, theory, and applications,''
\newblock {\em IEEE transactions on knowledge and data engineering}, vol. 35, no. 4, pp. 3313--3332, 2021.

\bibitem{ratliff2013characterization}
Lillian~J Ratliff, Samuel~A Burden, and S~Shankar Sastry,
\newblock ``Characterization and computation of local nash equilibria in continuous games,''
\newblock in {\em 2013 51st Annual Allerton Conference on Communication, Control, and Computing (Allerton)}. IEEE, 2013, pp. 917--924.

\bibitem{gulrajani2017improved}
Ishaan Gulrajani, Faruk Ahmed, Martin Arjovsky, Vincent Dumoulin, and Aaron~C Courville,
\newblock ``Improved training of wasserstein {GANs},''
\newblock {\em Advances in neural information processing systems}, vol. 30, 2017.

\end{thebibliography}
\end{document}